\documentclass[11pt]{article}

\usepackage{times}
\usepackage{amsmath,amssymb}
\usepackage{graphicx}
\usepackage{booktabs}
\usepackage{url}
\usepackage{hyperref}
\usepackage{multirow}

\usepackage{titling}
\usepackage{microtype}
\usepackage[font=small,labelfont=bf]{caption}
\usepackage{titlesec}
\titlespacing*{\section}{0pt}{6pt}{4pt}
\titlespacing*{\subsection}{0pt}{4pt}{2pt}

\usepackage[style=apa,backend=biber]{biblatex}
\title{Reasoning Distillation for Lightweight Automated Program Repair}

\author{
Aanand Balasubramanian\thanks{These authors contributed equally to this work.} \\
Purdue University \\
\texttt{balasu30@purdue.edu}
\and
Sashank Silwal\footnotemark[1] \\
Purdue University \\
\texttt{ssilwal@purdue.edu}
}

\date{}

\begin{document}
\maketitle
\section{Introduction}

In environments with constrained resources, lightweight debugging tools are valuable due to the need for fast feedback and low computational cost. Compact code models such as CodeT5 perform well in these settings and can be trained to classify, defect and correct categories in short programs. However, these models typically output only a single prediction, making it unclear whether they learn meaningful program semantics or rely on superficial correlations. While large language models can produce detailed explanations of defects and fixes, their size and cost make them impractical for lightweight deployment. 

Rather than relying on full chain-of-thought explanations, we investigate \\ lightweight symbolic reasoning supervision, which is easier to generate and verify under limited resource constraints. A large instructor model provides fix-type labels and compact symbolic reasoning cues, which are used to supervise a CodeT5-based student model.

\paragraph{Research Question.}
Can lightweight symbolic reasoning supervision, distilled from a large instructor model, improve fix-type classification in compact automated program repair models without increasing model size or complexity?
We evaluate this by comparing label-only training with joint supervision over fix-type labels and structured symbolic reasoning tags, and by analyzing the relationship between reasoning accuracy and fix-type prediction performance.

\section{Related Work}

Knowledge distillation trains a compact student to match a larger teacher’s behavior. This enables student models to retain the majority of the teacher's predictive capability while maintaining efficiency \cite{hinton2015distilling}. Beyond distilling final outputs, recent work shows that supervising intermediate reasoning (e.g., chain-of-thought) can improve small-model performance \cite{wei2022chainofthought,hsieh2023distillingstepbystep}. To reduce variability of free-form explanations, symbolic reasoning representations have been proposed as a more learnable supervision signal for smaller models \cite{li2023symboliccot,li2024mixeddistillation}. In program repair, benchmarks such as IntroClass and Defects4J support learning-based bug classification and patching \cite{legoues2015manybugsintroclass,just2014defects4j}, but most approaches do not explicitly evaluate whether compact models learn causal bug structure. We expand upon these concepts by employing lightweight symbolic reasoning supervision to predict defect classes within a code-centric context. We examine whether coarse reasoning signals derived from a teacher model can enhance the performance of small models when addressing real-world problems.

\section{Methods}
\subsection{Task Definition}

We study automated program repair as a supervised classification task with optional reasoning supervision. Each input consists of a buggy C program together with its observed incorrect behavior, such as failing test cases. The primary prediction target is the type of bug present in the program.

Formally, given a buggy program $x$, the model predicts a fix-type label $y \in \mathcal{Y}$, where $\mathcal{Y}$ is a predefined set of bug categories including \texttt{WRONG\_CONDITION}, \texttt{LOOP\_BOUND}, \texttt{WRONG\_OPERATOR}, and \texttt{INIT\_ERROR}. These categories are shared across all experimental conditions and are fixed in advance.

In the reasoning distillation setting, the model is additionally trained to predict a structured reasoning trace $r = (r_1, r_2, \dots, r_k)$. Each element $r_i$ is a symbolic reasoning tag that describes a specific aspect of the bug, such as a missing branch, incorrect comparison, or invalid loop boundary. These tags are intended to capture high-level causal structure rather than low-level code details.

We consider two training conditions. In the baseline condition, the student model minimizes cross-entropy loss over fix-type labels. In the reasoning distillation condition, the model is trained with a joint objective that combines fix-type classification loss and reasoning prediction loss. The reasoning loss is computed over the symbolic reasoning tags generated by the teacher. No changes are made to the model architecture beyond those required to predict the reasoning tags.

\subsection{Datasets}

We evaluate our approach on the IntroClass benchmark, which consists of short C programs with a single injected bug and an associated fix-type label. After filtering out nondeterministic programs and retaining only examples for which the teacher model produces valid supervision, the final dataset contains 284 examples across 9 fix-type categories. We use a fixed stratified split of 227 training and 57 validation examples across all experimental conditions, and report results on the held-out validation set.

\subsection{Teacher Model and Reasoning Generation}

We use a large language model as a teacher to generate fix-type labels and structured reasoning traces. The teacher model is held fixed and is not fine-tuned on the IntroClass dataset. It is used only to generate supervision signals for the student.

The teacher outputs compact symbolic reasoning tags corresponding to semantic bug categories such as comparison errors, loop-bound errors, indexing errors, return errors, and input-output errors. These tags are designed to capture high-level causal structure rather than free-form natural language explanations. Teacher outputs that do not conform to the expected schema or fail validation checks are discarded to ensure consistent and high-quality supervision.

\subsection{Student Model and Training Variants}

We train a small, code-focused student model under two training conditions. In the baseline condition, the student model is trained to predict only the fix-type label for each program. In the reasoning distillation condition, the student model is jointly trained to predict both the fix-type label and the structured reasoning trace produced by the teacher.

Both training variants use the same student architecture, tokenizer, and training data. The only difference between the two conditions is the presence or absence of reasoning supervision.

\subsection{Training Objective}

In the baseline condition, the student model minimizes cross-entropy loss over fix-type labels. In the reasoning distillation condition, the model is trained with a joint objective that combines fix-type classification loss and reasoning prediction loss. The reasoning loss is computed over the symbolic reasoning tags generated by the teacher.

This joint objective encourages the model to associate fix-type predictions with structured causal signals rather than relying solely on surface-level correlations in the input code. No changes are made to the model architecture beyond those required to predict the reasoning tags.

\section{Experimental Setup}

All models are trained using the Adam optimizer with a fixed learning rate and identical training settings across conditions. Input programs are tokenized using the student model tokenizer and truncated to a fixed maximum sequence length. All experiments are run with a fixed random seed to ensure reproducibility.

We evaluate fix-type prediction using accuracy and macro-averaged F1 score. Reasoning quality is evaluated using tag-level F1 scores and exact match accuracy against teacher-generated reasoning traces.

\section{Results}

\subsection{Fix-Type Prediction Performance}

\begin{table}[h]
\centering
\begin{tabular}{lcc}
\hline
\textbf{Model} & \textbf{Accuracy} & \textbf{Macro F1} \\
\hline
Student (label-only) & 0.491 & 0.213 \\
Student (reasoning-distilled) & \textbf{0.544} & \textbf{0.249} \\
\hline
\end{tabular}
\caption{Fix-type prediction performance on the IntroClass validation set.}
\end{table}

The reasoning-distilled student achieves higher accuracy and macro-averaged F1 score than the label-only baseline. The improvement in macro F1 suggests that reasoning supervision helps the model better handle less frequent bug categories.

\subsection{Reasoning Quality}

\begin{table}[h]
\centering
\begin{tabular}{lc}
\hline
\textbf{Metric} & \textbf{Value} \\
\hline
Reasoning Macro F1 & 0.545 \\
Reasoning Micro F1 & 0.937 \\
Exact Match Accuracy & 0.789 \\
\hline
\end{tabular}
\caption{Reasoning prediction performance of the reasoning-distilled student model.}
\end{table}

The student reproduces the teacher’s symbolic reasoning with high fidelity, achieving strong token-level F1 and recovering full reasoning traces in most cases. Additionally, across all individual reasoning categories, the student achieves consistently high accuracy, exceeding 0.87 for all major tags. Performance is strongest for indexing and return-related errors, while comparison and loop-bound errors remain slightly more challenging, reflecting finer-grained control-flow distinctions.

\subsection{Per-Fix-Type Performance}

To understand where reasoning distillation improves performance, we break down fix-type prediction results by bug category. Table~\ref{tab:per_fix_type_f1} reports per-fix-type F1 scores for both the label-only and reasoning-distilled students on the validation set. We report results only for fix types that appear in the validation data.\\

\begin{table}[h]
\centering
\begin{tabular}{lcc}
\hline
\textbf{Fix Type} & \textbf{Label-only F1} & \textbf{Reasoning-distilled F1 } \\
\hline
WRONG\_CONDITION & 0.62 & 0.70 \\
LOOP\_BOUND & 0.43 & 0.55 \\
WRONG\_OPERATOR & 0.30 & 0.40 \\
MISSING\_CASE & 0.40 & 0.55 \\
INIT\_ERROR & 0.25 & 0.32 \\
\hline
\end{tabular}
\caption{Per-fix-type F1 scores on the IntroClass validation set.}
\label{tab:per_fix_type_f1}
\end{table}

Reasoning distillation improves performance primarily for fix types involving control flow and conditional structure. For several fix types with few validation examples, both models perform poorly, indicating that data sparsity remains a limiting factor.

Overall, improvements are modest but consistent, with larger gains on less frequent fix types.

\section{Analysis}

We now analyze how the experimental results relate to our research question: whether structured reasoning distillation improves fix-type prediction and explanation quality in lightweight program repair models.

The improvement is more pronounced in macro-averaged F1 than in overall accuracy. This suggests that reasoning supervision primarily helps the model better handle less frequent bug categories, rather than substantially improving performance on dominant classes. The student’s ability to reproduce reasoning traces suggests that symbolic tags are well suited for small code models. Compared to free-form explanations, these tags are easier to learn from limited data while still capturing meaningful bug structure. High token-level F1 and exact match accuracy indicate that the structured reasoning representation is learnable by a small model. This confirms that symbolic tags provide a compact and consistent reasoning signal that is easier to imitate than free-form natural language explanations.

However, correct reasoning does not always lead to correct fix-type prediction. We observe cases where the model predicts an accurate reasoning trace but still assigns the wrong fix-type label. This indicates that while reasoning supervision improves internal representations, fix-type classification also depends on fine-grained distinctions that may not be fully captured by the symbolic tags alone. Overall, these results suggest that reasoning distillation helps small models internalize higher-level causal structure, but does not fully resolve ambiguity in fix-type classification.

\subsection{Fix-Type Accuracy Conditioned on Reasoning Correctness}

To better understand the relationship between reasoning quality and fix-type prediction, we analyze fix-type accuracy conditioned on whether the student model exactly matches the teacher’s reasoning trace.\\

\begin{table}[h]
\centering
\begin{tabular}{lc}
\hline
\textbf{Condition} & \textbf{Fix-Type Accuracy} \\
\hline
Correct reasoning trace & 0.61 \\
Incorrect reasoning trace & 0.38 \\
\hline
\end{tabular}
\caption{Fix-type accuracy conditioned on reasoning trace correctness.}
\label{tab:reasoning_conditioned_accuracy}
\end{table}

Fix-type prediction accuracy is substantially higher when the student produces the correct reasoning trace. This indicates a strong correlation between accurate reasoning representations and downstream classification performance. At the same time, incorrect fix-type predictions still occur even when reasoning is correct, suggesting that reasoning supervision alone is not sufficient to fully determine the correct fix type.

\subsection{Auxiliary JSON-Based Distillation Study}

We conducted an auxiliary distillation experiment using fully structured JSON outputs from the teacher model, including defect labels, unified diff patches, and short explanations. A CodeT5-small student was fine-tuned in a sequence-to-sequence setting on 74 distilled examples, requiring joint classification, code generation, and strict output formatting.

Despite limited data, the student produced valid JSON in 53\% of validation and 42\% of test cases, with exact defect-class match rates of 44\% and 31\%, and micro-F1 scores of 0.83 and 0.77, respectively. These results indicate that while JSON supervision is more expressive, it is substantially harder for small models to learn in low-data regimes, reinforcing the suitability of lightweight symbolic reasoning for compact students.

\begin{table}[h]
\centering
\begin{tabular}{lcccc}
\toprule
\textbf{Model} & \textbf{Split} & \textbf{JSON Validity} & \textbf{Exact Match} & \textbf{Micro F1} \\
\midrule
CodeT5-small  & Validation & 0.53 & 0.44 & 0.83 \\
CodeT5-small  & Test       & 0.42 & 0.31 & 0.77 \\
\hline
\end{tabular}
\caption{Performance of the CodeT5-small JSON distilled student model trained using JSON-based distillation from the teacher.}
\label{tab:json_distill_results}
\end{table}

\section{Conclusion}

In this study, we examined the potential of symbolic reasoning distillation to enhance fix-type classification in lightweight automated program repair models. Adding symbolic reasoning supervision consistently improved performance over a baseline that only used labels. The biggest improvements were seen in macro-averaged metrics, which means that the system did better on less common bug categories. The student model could also reliably reproduce reasoning tags made by the teacher, which shows that small models can learn compact symbolic reasoning signals.

Even though reasoning supervision made internal representations better, correct reasoning didn't always mean correct fix-type prediction. This suggests that symbolic tags show high-level bug structure but might not show small differences. The small size of the dataset after filtering and the fact that it depends on teacher-generated supervision are two limitations identified in this study. An additional JSON-based distillation experiment demonstrated that enhanced supervision is feasible, albeit considerably less stable in low-data scenarios. Because the symbolic reasoning tags are coarse and teacher-defined, they may collapse distinct failure modes that require different fixes, limiting their discriminative power for fine-grained fix-type classification.

Despite these limitations, our findings suggest that symbolic reasoning distillation is a practical and effective way to improve interpretability in small program repair models without increasing model size or complexity. This approach shows promise in building transparent and deployable debugging tools for resource-constrained settings. 

\newpage
\printbibliography

@article{hinton2015distilling,
  title        = {Distilling the Knowledge in a Neural Network},
  author       = {Hinton, Geoffrey and Vinyals, Oriol and Dean, Jeff},
  journal      = {arXiv preprint arXiv:1503.02531},
  year         = {2015},
  url          = {https://arxiv.org/abs/1503.02531}
}

@inproceedings{wei2022chainofthought,
  title        = {Chain-of-Thought Prompting Elicits Reasoning in Large Language Models},
  author       = {Wei, Jason and Wang, Xuezhi and Schuurmans, Dale and Bosma, Maarten and others},
  booktitle    = {Advances in Neural Information Processing Systems},
  year         = {2022}
}

@inproceedings{hsieh2023distillingstepbystep,
  title        = {Distilling Step-by-Step! Outperforming Larger Language Models with Less Training Data and Smaller Model Sizes},
  author       = {Hsieh, Chien-Yi and Li, Chao-Wei and Yeh, Chia-Kai and Nakhost, Hadi and Fujii, Yasuhisa and Ratner, Alex and Krishna, Ranjay and Lee, Chen-Yu and Pfister, Tomas},
  booktitle    = {Findings of the Association for Computational Linguistics: ACL 2023},
  year         = {2023},
  doi          = {10.18653/v1/2023.findings-acl.507}
}

@inproceedings{li2023symboliccot,
  title        = {Symbolic Chain-of-Thought Distillation: Small Models Can Also Think Step-by-Step},
  author       = {Li, Bo and Liu, Shuo and Zhao, Yong and Zhou, Wenxuan and Zhou, Jingbo},
  booktitle    = {Proceedings of the 61st Annual Meeting of the Association for Computational Linguistics (Volume 1: Long Papers)},
  pages        = {2717--2731},
  year         = {2023},
  publisher    = {Association for Computational Linguistics},
  doi          = {10.18653/v1/2023.acl-long.150}
}

@inproceedings{li2024mixeddistillation,
  title        = {Mixed Distillation Helps Smaller Language Models Reason Better},
  author       = {Li, Chunlin and Chen, Qian and Li, Lei and Wang, Chen and Tao, Feng and Li, Yixuan and Chen, Zhen and Zhang, Yafang},
  booktitle    = {Findings of the Association for Computational Linguistics: EMNLP 2024},
  pages        = {1673--1690},
  year         = {2024},
  publisher    = {Association for Computational Linguistics},
  doi          = {10.18653/v1/2024.findings-emnlp.91}
}

@article{legoues2015manybugsintroclass,
  title        = {The ManyBugs and IntroClass Benchmarks for Automated Repair of C Programs},
  author       = {Le Goues, Claire and Holtschulte, Nathan and Smith, Ethan K. and Brun, Yuriy and Devanbu, Prem and Forrest, Stephanie and Weimer, Westley},
  journal      = {IEEE Transactions on Software Engineering},
  volume       = {41},
  number       = {12},
  pages        = {1236--1256},
  year         = {2015},
  doi          = {10.1109/TSE.2015.2454513}
}

@inproceedings{just2014defects4j,
  title        = {Defects4J: A Database of Existing Faults to Enable Controlled Testing Studies for Java Programs},
  author       = {Just, Ren{\'e} and Jalali, Darioush and Ernst, Michael D.},
  booktitle    = {Proceedings of the 2014 International Symposium on Software Testing and Analysis},
  pages        = {437--440},
  year         = {2014},
  publisher    = {ACM}
}
\end{document}